\documentclass[10pt,twocolumn,letterpaper]{article}

\usepackage{cvpr}              

\usepackage{amsmath,amsfonts,bm}
\usepackage{algorithm,algorithmic}

\def\eqref#1{equation~\ref{#1}}
\def\Eqref#1{Equation~\ref{#1}}

\def\1{\bm{1}}

\def\RVec#1{{\mathbf{#1}}}

\def\RMat#1{{\mathbf{#1}}}

\DeclareMathAlphabet{\mathsfit}{\encodingdefault}{\sfdefault}{m}{sl}
\SetMathAlphabet{\mathsfit}{bold}{\encodingdefault}{\sfdefault}{bx}{n}

\def\Set#1{{\mathbb{#1}}}

\usepackage[dvipsnames]{xcolor}

\definecolor{cvprblue}{rgb}{0.21,0.49,0.74}
\usepackage[pagebackref,breaklinks,colorlinks,citecolor=cvprblue]{hyperref}
\usepackage{arydshln}
\usepackage{multirow}

\def\paperID{13} 
\def\confName{CVPR}
\def\confYear{2024}

\title{Latent-based Diffusion Model for Long-tailed Recognition}

\author{Pengxiao Han$^{1}$, 
Changkun Ye$^{1,2}$, 
Jieming Zhou$^{1,2}$,
Jing Zhang$^{1}$,
Jie Hong$^{1,2}$,
Xuesong Li$^{1,2}$\thanks{Corresponding author}  \\
$^{1}$Australian National University, 
$^{2}$CSIRO, Australia \\ 
{\tt\small \{pengxiao.han, changkun.ye, jieming.zhou, jing.zhang, jie.hong, xuesong.li\}@anu.edu.au} \\
}

\begin{document}
\maketitle
\begin{abstract}
Long-tailed imbalance distribution is a common issue in practical computer vision applications. Previous works proposed methods to address this problem, which can be categorized into several classes: re-sampling, re-weighting, transfer learning, and feature augmentation. In recent years, diffusion models have shown an impressive generation ability in many sub-problems of deep computer vision. However, its powerful generation has not been explored in long-tailed problems. We propose a new approach, the Latent-based Diffusion Model for Long-tailed Recognition (LDMLR), as a feature augmentation method to tackle the issue. First, we encode the imbalanced dataset into features using the baseline model. Then, we train a Denoising Diffusion Implicit Model (DDIM) using these encoded features to generate pseudo-features. Finally, we train the classifier using the encoded and pseudo-features from the previous two steps. The model's accuracy shows an improvement on the CIFAR-LT and ImageNet-LT datasets by using the proposed method. Code is available for research at \href{https://github.com/AlvinHan123/LDMLR}{https://github.com/AlvinHan123/LDMLR}
\end{abstract}    
\section{Introduction}
Long-tailed recognition is a crucial task in deep computer vision because the imbalanced data distributions are close to real-world applications~\cite{drosou2014support, makki2019experimental, li2019detection}.  In many real-world datasets which have limited-labelled data, some classes of data have many samples, while others have few. In healthcare, many disease instances follow a long-tailed distribution~\cite{drosou2014support, zeng2016effective, yang2022proco, holste2022long}. As for fraud detection~\cite{makki2019experimental, shamsudin2020combining, zhu2020optimizing}, the number of samples of fraudulent transactions is much smaller than that of legitimate ones. Addressing long-tailed recognition could enhance the robustness of deep-learning visual models on real-world applications. However, long-tailed recognition is challenging since deep neural networks are more prone to overfitting the majority class while damaging the prediction accuracy for minority classes.

So far, extensive research has been conducted to address dataset imbalance issues~\cite {ouyang2016factors, zhou2005training, lin2017focal, cao2019learning, han2023wrapped}. Class-sensitive learning effectively solves the long-tailed distribution problem~\cite{elkan2001foundations, zhou2005training, lin2017focal, cao2019learning}. Class-sensitive learning addresses this issue by re-adjusting the training loss for different classes. Logit adjustment methods adjust the prediction logits based on label frequencies~\cite{menon2020long, tian2020posterior_labelshift, zhang2021distribution, kini2021label, hong2021disentangling}. However, class-sensitive learning and logit adjustment methods rely too heavily on training label frequencies. There are also representation learning methods, such as prototype learning~\cite{liu2019large, zhong2019unequal}, metric learning~\cite{hermans2017defense, zhang2017range}, and sequential learning~\cite{ouyang2016factors, zhong2019unequal}. These methods use the feature representation of each class to make the distinction between classes more significant. Although these methods can effectively improve the accuracy of deep learning networks on long-tailed distributed datasets, they often need to be carefully designed and have limited improvements in tail classes.

Data or feature augmentation is another effective solution in long-tailed recognition~\cite{yin2019feature, shelke2017review}. Some augmentation methods attempt to transfer the knowledge from head classes to tail to augment the samples of tail classes~\cite{yin2019feature, kim2020m2m, chen2023transfer}. Others use over-sampling and under-sampling to re-balance the datasets~\cite{shelke2017review}. Additionally, it is natural to come that using a high-quality generative model to augment a long-tailed distributed dataset might improve the performance. The generative models have been well developed~\cite{goodfellow2014generative, sohn2015learning, ho2020denoising_DDPM}, and variational autoencoder (VAE) and generative adversarial network (GAN) have been applied in the long-tailed problems~\cite{ali2019mfc, yang2021ida}. In recent years, diffusion model series have shown superior ability than other generative models~\cite{ho2020denoising_DDPM, nichol2021improved_IDDPM, song2022denoising_DDIM, nokey_ldm}. However, the application of such powerful generative models is underexplored in long-tailed problems.

Inspired by the observation above, our work attempts to leverage the diffusion model for feature augmentation to address long-tailed distribution recognition. We propose the approach named Latent-based Diffusion Model for Long-tailed Recognition (LDMLR). Specifically, we first train a baseline on the long-tailed data and obtain the encoded features. Then, we use a Denoising Diffusion Implicit Model (DDIM) model to generate pseudo-features in the latent space to augment the long-tailed training dataset. The augmentation in latent space reduces the computational cost and speeds up the training process. Finally, we use both the encoded and pseudo-features to train the classifier of LDMLR to predict the long-tailed data. The proposed LDMLR has been validated on CIFAR-LT~\cite{liu2019large, krizhevsky2009learning} and ImageNet-LT~\cite{liu2019large, ILSVRC15}. The experimental results demonstrate that our method is beneficial for long-tailed recognition. Our contribution can be summarized as below:
\begin{itemize}
    \item Our method applies the diffusion model to enrich the feature embeddings for the long-tailed problem, offering a new solution to this challenging problem. To the best of my knowledge, we are first to explore the capability of diffusion model in the long-tailed recognition problem.
    \item When using the diffusion model, we propose to do the augmentation in the latent space instead of the image space, which reduces the computational cost and speeds up the training process.
    \item The experiments demonstrate that LDMLR has improved the performance of long-tailed recognition tasks on different datasets using various baselines. We achieve the essential improvements over the baselines.
\end{itemize}
\section{Related Works}
\subsection{Long-tailed Recognition}
Due to the class imbalance within datasets, long-tailed recognition is a challenging problem. A neural network trained on long-tailed datasets is prone to be biased towards head (or majority) classes with enough training data, resulting in poor performance on tail  (or minority) classes. Class-sensitive learning aims to address the class imbalance problem by readjusting the traditional softmax cross-entropy loss. Traditional cross-entropy tends to provide more gradients to head classes, while tail classes receive fewer gradients. To ensure that each class has a balanced impact on the neural network during training, class-sensitive learning proposes adjusting the training loss weights for each class based on given training label frequencies, such as Focal loss~\cite{lin2017focal} and Label-distribution-aware-margin (LDAM) loss~\cite{cao2019learning}. 

Logit adjustment is another method for addressing the recognition of imbalanced datasets. This approach shares similarities with class-sensitive learning, where most of both methods require training class frequencies to rebalance the influence of head and tail classes on the model. For example, \cite{menon2020long} modifies logit adjustment based on label frequencies, which can be implemented by posthoc or enforcement of a large relative margin between the logits of tail versus head labels. Label Shift Compensation (LSC) \cite{tian2020posterior_labelshift, zhang2021distribution, Ye_2024_WACV_labelshift} is another type of logit adjustment method. Most logit adjustment methods require training label frequencies, even when not required. Some logit adjustment methods introduce an additional model, which makes the entire process slower and more complex. Representation learning involves bringing images into the feature space to learn discrimination classes. WCDAS~\cite{han2023wrapped} is a representation learning-based method that incorporates data-wise Gaussian-based kernels into the angular correlation between feature representation and classifier weights. In addition to class-sensitive learning and logit adjustment, there is the method that utilize knowledge distillation to address the long-tailed recognition problem like Self-Supervision to Distillation (SSD)~\cite{li2021self}.

\subsection{Generative Models for Feature Augmentation in Long-tailed Recognition}
Many recent works demonstrate that feature or sample augmentation can effectively address the long-tailed recognition problem~\cite{ali2019mfc, yang2021ida, zang2021fasa, hong2022safa}. By enriching the features or samples for the tail class, this approach is highly compatible and can easily be combined with normal baseline models. Using a powerful generative model to do augmentation can effectively diversify an imbalanced dataset~\cite{ali2019mfc, yang2021ida}. MFC-GAN~\cite{ali2019mfc} proposes a conditional GAN with fake class labels to generate a small number of minority class instances, thereby re-balancing the entire dataset. This method performs data augmentation at the image level, which is feasible to apply to large-scale image datasets. IDA-GAN~\cite{yang2021ida} also uses GANs 
for data augmentation. First, they train a VAE encoder to model the training set distribution in the latent space. Then, they use GANs to generate tail class images to re-balance the long-tailed distributed dataset. 

It is known that the diffusion models~\cite{ho2020denoising_DDPM, nichol2021improved_IDDPM, song2022denoising_DDIM, nokey_ldm} are relatively underexplored in long-tailed recognition. Since the diffusion model is a probabilistic generative model, it can generate more diverse samples than GANs, which might be more effective for solving long-tailed recognition. Moreover, it has the better ability in generating high-quality samples than other generative methods. As such, in this paper, we propose a diffusion-based augmentation method, LDMLR. We hope to exploit the excellent ability of the diffusion model for the long-tailed problem.

\subsection{Diffusion Model}
Due to the remarkable generative results of diffusion models, they have been a recent emerging topic in computer vision \cite{ho2020denoising_DDPM, song2022denoising_DDIM, nichol2021improved_IDDPM, nokey_ldm, peebles2023scalable_dit, ramesh2022hierarchical_dalle2, saharia2022photorealistic_imagen}. As a likelihood-based generative model, a diffusion model can produce exact likelihood computation, showing a powerful generation ability. DDIM~\cite{song2022denoising_DDIM} removes the Markov chain constraint from Denoising Diffusion Probabilistic Models (DDPM)~\cite{ho2020denoising_DDPM}, which allows fewer steps to accelerate the sampling process. DDIM significantly speeds up the sampling process without damaging the generative quality. Despite DDIM's acceleration of the sampling speed, the training speed of a diffusion model is still considerably slow, which prevents its widespread adoption for large-scale images. The Latent Diffusion Model (LDM)~\cite{nokey_ldm} perfectly addresses this issue. LDM uses a VAE encoder to compress large-scale images into latent features that sequentially are used as a feature dataset for training a diffusion model in latent space. 

Our goal in using data augmentation for the long-tailed recognition task is to generate diverse samples, which might enrich the low-density region of the category distribution. In order to reduce the model complexity and the training time, we choose to augment features in the latent space. It is natural to use LDM. However, LDM is specifically designed for large-scale, high-resolution images. In our method, the diffusion model is used for generating low-dimensional features. Therefore, we propose to modify DDIM for augmenting features in the latent space.
\begin{figure*}[t]
  \centering
    \includegraphics[width=1.0\linewidth]{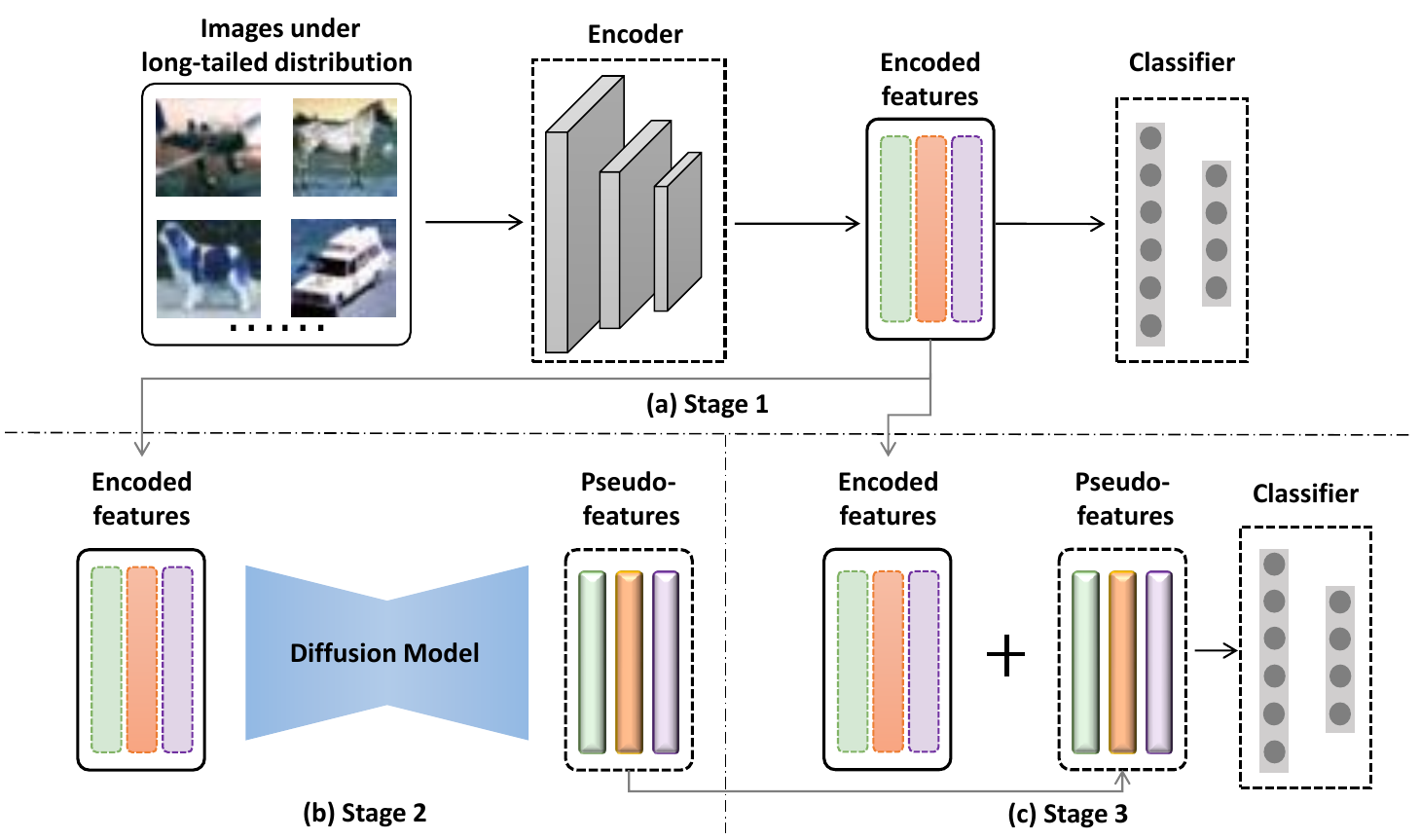}
    \caption{Overview of the proposed framework, LDMLR. The figure describes the training of the framework: (a) obtain encoded features by a pre-training convolutional neural network on the long-tailed training set, (b) Generate pseudo-features by the diffusion model using encoded features, and (c) Train the fully connected layers using encoded and pseudo-features. The encoder from (a) and the classifier from (c) are used to predict long-tailed data in the evaluation stage.}\label{framework}
\end{figure*}

\section{Approach}
We propose a three-stage model called LDMLR for the long-tailed classification problem, as shown in Figure~\ref{framework}. We first train a neural network model on the long-tailed dataset and extract ground truth encoded features of the ground truth images. A diffusion model is then trained to generate pseudo-features of each class. Finally, the classifier is fine-tuned with the encoded and pseudo-features. The algorithm is also described in Algorithm~\ref{alg:all}.

\subsection{Preliminaries}
\textbf{Notations:} Let $\mathcal{X}\subseteq \Set{R}^{d}$ be the image space, $\mathcal{Y}=\{1,2,...,K\}$ be the label space and $\mathcal{Z}\subseteq \Set{R}^c$ be the feature space. In the training time, a long-tailed dataset $\mathcal{D}=\{(x_i,y_i)\}^N_{i=1}$ is available, with $(x_i,y_i)\sim_{i.i.d} P(x,y)$ drawn from the data distribution $P(x,y)$.

\noindent\textbf{Diffusion Model:}
The diffusion model is a type of generative model. A typical diffusion model consists of two stages -- a forward process that gradually adds noise to a clean sample and a backward process that gradually recovers a clean sample from noise. This paper mainly considers a special type of diffusion model named DDIM~\cite{song2022denoising_DDIM}. 

In the forward process, given a clean sample $x_0\sim q(x_0)$, a series of noise samples $x_1, x_2,..., x_T$ are generated by:
\begin{equation}\label{eq:diff-forward}
    q(x_t|x_{t-1}) := \mathcal{N}(x_t;\sqrt{\alpha_t} \cdot x_t,(1-\alpha_t) \mathbf{I}),
\end{equation}
for $\forall t\in \{1,2,...,T\}$, where $T$ is the number of diffusion steps, $1-\alpha_t$ is the noise variance at step $t$ and $\mathbf{I}$ is the identity matrix with the same dimension as $x_0$. It is expected that $q(x_T|x_0)\approx\mathcal{N}(0,\mathbf{I})$ is close to the Gaussian noise.

In the backward process, starting from $x_T\sim\mathcal{N}(0,\mathbf{I})$, DDIM aims to recover the clean sample $x_0$ and $x_{\tau_1},...,x_{\tau_s}$, where $\{\tau_1,...,\tau_s\}\subseteq \{1,2,...,T\}$ is a subset of the forward steps. The backward process satisfies:
\begin{equation}\label{eq:diff-backward}
    p(x_{t-1}|x_t, x_0) = \mathcal{N}(x_{t-1};\mu(x_t, x_0, t),\Sigma(x_t, x_0, t)),  
\end{equation}
where:
\begin{align}
\mu(x_t, x_0, t)     &= \sqrt{\alpha_{t-1}}x_0 + \sqrt{1 - \alpha_{t-1} - \sigma_t^2}\cdot \frac{x_t-\sqrt{\alpha_t}x_0}{\sqrt{1-\alpha_t}}, \\
\Sigma(x_t, x_0, t)  &= \sigma_t^2 \mathbf{I}.
\end{align}

While $\mu_\theta (x_t,x_0)$ can be modeled directly by a Neural Network, a recent study suggests that modeling the noise term $\epsilon_\theta(x_t,t)$ instead leads to better performance \cite{ho2020denoising_DDPM}. In this case, the backward step then satisfies:
\begin{equation}\label{eq:diff-backward-approx}
\begin{aligned}
    x_{t-1} = & \sqrt{\alpha_{t-1}} \left(\frac{x_t-\sqrt{1-\alpha_t}\cdot \epsilon_\theta(x_t,t)}{\sqrt{\alpha_t}}\right) \\
    & +  \sqrt{1 - \alpha_{t-1} - \sigma_t^2}\cdot \epsilon_\theta(x_t,t) + \sigma_t\epsilon_t,
\end{aligned}
\end{equation}
where $\epsilon_t\sim\mathcal{N}(0,\mathbf{I})$ is an independent random noise.

After creating noise samples $x_1,x_2,...,x_T$ in the forward process and learning $\epsilon_\theta (x_t,t)$ in the backward process, the diffusion model is able to generate data samples from Gaussian noises based on \Eqref{eq:diff-backward-approx}.

\subsection{Stage 1: Image Encoding}
In order to augment image features with a latent diffusion model, the first step of LDMLR is to learn a neural network feature extractor $\mathcal{E}:\mathcal{X}\rightarrow\mathcal{Z}$ and obtain decent feature representations of the images in the dataset $\mathcal{D}$ (see Figure~\ref{framework} (a)). 

Since $\mathcal{D}$ is a labelled dataset, we propose to construct a soft classifier $f:\mathcal{X}\rightarrow\Delta^{K-1}$ based on the feature extractor $\mathcal{E}$ by adding a classification head $\mathcal{G}:\mathcal{Z}\rightarrow\Delta^{K-1}$ over $\mathcal{E}$:
\begin{equation}
    f(x) = \mathcal{G}(\mathcal{E}(x)),
\end{equation}
where $\Delta^{K-1}$ is the space of softmax predictions.

The feature extractor $\mathcal{E}$ can then be jointly trained with $\mathcal{G}$ over the dataset $\mathcal{D}$ with a cross entropy loss:
\begin{equation}
    \mathcal{L}_{CE} := -\mathbb{E}_{(x,y)}\log f(x)_{y},
\end{equation}
where $(x,y)\sim p(x,y)$ follows train set distribution.

After training the classifier $f$, the encoder $\mathcal{E}$ can output features using ground truth images from the dataset $\mathcal{D}$. The set of the labelled encoded features is denoted as:
\begin{equation}
    \mathcal{D}_z = \{(z_i,y_i)|z_i = \mathcal{E}(x_i), (x_i,y_i)\in\mathcal{D}\}.
\end{equation}

With $\mathcal{D}_z$ available, we can train the latent diffusion model in the next stage to generate pseudo-features.

\subsection{Stage 2: Representation Generation}
As shown in Figure~\ref{framework} (b), in the second stage of LDMLR, we propose to train a class-conditional latent diffusion model (LDM) to generate pseudo-feature representations for different classes. The latent diffusion model has several advantages over image generation diffusion models: 1) LDM operates on one-dimensional image features instead of the original high-dimensional images. Therefore, LDM is more efficient than the image generation diffusion model in terms of training and inference time; 2) Standard diffusion model trained on a long-tailed dataset can suffer from low diversity and fidelity problems due to insufficient images in tail classes~\cite{Qin_2023_CVPR, zhang2024longtailed}, while LDM may suffer less from this problem because the data has lower dimensionality.

In the proposed LDMLR, we adopt the DDIM~\cite{song2022denoising_DDIM} diffusion model approach to train the model with the encoded $\mathcal{D}_{z}$ and generate pseudo-features. In the forward process, following \Eqref{eq:diff-forward}, an encoded feature $z_0\in\mathcal{D}_{z}$ is perturbed with Gaussian noise to create $z_1,...,z_T$ with:
\begin{equation}
    q(z_t|z_{t-1}) := \mathcal{N}(z_t;\sqrt{\alpha_t} \cdot z_t,(1-\alpha_t) \mathbf{I}).
\end{equation}

In the backward process, we propose to train a class-conditional neural network model $\epsilon_\theta(z_t,t,y)$ to approximate the Gaussian noise $\epsilon(z_t,t)$. The input of the neural network model includes the noisy feature $z_t$ and the condition embedding determined by the step $t$ and label $y\in\mathcal{Y}$.

The neural network model is trained with an MSE loss that minimizes the L2 distance between the predicted noise and the ground truth noise, which is defined as:
\begin{equation}
\mathcal{L}_{LDM} := \mathbb{E}_{z, y, \epsilon, \tau}\left[\| \epsilon - \epsilon_\theta(z_t, t, y) \|_2^2\right],
\end{equation}
where $z=\mathcal{E}(x), (x,y)\sim p(x,y), \epsilon\sim\mathcal{N}(0,\mathbf{I})$ and $t$ is drawn uniformly from $\{1, 2,...,T\}$.

After training the LDM, we can use $\epsilon_\theta(z_t, t, y)$ to generate pseudo-feature representations from $z_T\sim\mathcal{N}(0,\mathbf{I})$ for each class $y\in\mathcal{Y}$ based on the \Eqref{eq:diff-backward-approx}:
\begin{equation}\label{eq:diff-backward-approx-p}
\begin{aligned}
    \hat{z}_{t-1} = & \sqrt{\alpha_{t-1}} \left(\frac{\hat{z}_t-\sqrt{1-\alpha_t}\cdot \epsilon_\theta(\hat{z_t},t,y)}{\sqrt{\alpha_t}}\right) \\
    & +  \sqrt{1 - \alpha_{t-1} - \sigma_t^2}\cdot \epsilon_\theta(\hat{z_t},t,y) + \sigma_t\epsilon_t,
\end{aligned}
\end{equation}

These labelled pseudo-features are then collected as:
\begin{equation}
    \mathcal{D}_{\hat{z}} = \{(\hat{z_i},y_i)| y_i\sim p(y),\hat{z_i} \text{ generated by LDM} \}
\end{equation}
where $p(y)$ can be different distributions on demand. For example, we can choose $p(y)$ to be none zero on tail classes so that the LDM only generates tail class features.

\subsection{Stage 3: Classifier Training}
In the final stage of LDMLR (see Figure~\ref{framework} (c)), we fine-tune the classification head $\mathcal{G}$ with labelled encoded feature $\mathcal{D}_z$ obtained from stage 1 and labelled pseudo-feature $\mathcal{D}_{\hat{z}}$ obtained in stage 2 by the latent diffusion model.

\begin{equation}\label{eq:loss-stage-3}
    \mathcal{L}_{FT} := -\mathbb{E}_{(z,y)}\left[\log \mathcal{G}(z)_{y}\right] - \gamma \mathbb{E}_{(\hat{z},y)}\left[\log \mathcal{G}(\hat{z})_{y}\right],
\end{equation}
where $z = \mathcal{E}(x), (x,y)\sim p(x,y)$ and $(\hat{z},y)\in\mathcal{D}_{\hat{z}}$, and $\gamma$ is the hyperparameter that determines the relative contribution of the two terms in the loss.

After fine-tuning the classification head $\mathcal{G}$, we can combine the feature extractor $\mathcal{E}$ and $\mathcal{G}$ to construct the final classifier $f(x) = \mathcal{G}(\mathcal{E}(x))$ for the long-tailed classification problem. The general structure of our model has been summarised in Algorithm \ref{alg:all}.

\begin{algorithm}[h]
\caption{Proposed LDMLR}
	\begin{algorithmic}
	    \label{alg:all}
	    \STATE \textbf{Input: } Data $\mathcal{D}$, Model $\mathcal{E}(x), \mathcal{G}(z), \epsilon_\theta(z_t,t,y)$. 
        \STATE \textbf{Stage 1:} Image Encoding:
                \begin{itemize}
                    \item Train $f(x)=\mathcal{G}(\mathcal{E})(x)$ with $\mathcal{D}$ and loss $\mathcal{L}_{CE}$.
                    \item Extract labelled features $\mathcal{D}_{z}$.
                \end{itemize}
        \STATE \textbf{Stage 2:} Representation Generation
                \begin{itemize}
                    \item Train $\epsilon_\theta(z_t,t,y)$ with $\mathcal{D}_{z}$ and loss $\mathcal{L}_{LDM}$.
                    \item Generate labelled pseudo-features $\mathcal{D}_{\hat{z}}$.
                \end{itemize}
        \STATE \textbf{Stage 3:} Classifier Training
                \begin{itemize}
                    \item Fine-tuning $\mathcal{G}$ with $\mathcal{D}_{z} + \mathcal{D}_{\hat{z}}$ and loss $\mathcal{L}_{FT}$.
                \end{itemize}
		\STATE \textbf{Output: } Classifier $f(x)=\mathcal{G}(\mathcal{E}(x))$.
	\end{algorithmic} 
\end{algorithm}  
\section{Experiments}
\begin{table*}[t]
\caption{Experimental results on CIFAR-LT~\cite{liu2019large}. The classification accuracies in $\%$ are provided. ``$\uparrow$'' indicates the improvements over the baseline. The best numbers are in bold. The results of CE, Label Shift, and WCDAS are obtained by self-implemented networks. FASA~\cite{zang2021fasa} and SAFA~\cite{hong2022safa} are feature-augmentation-based methods.}
\label{tab:CIFAR_LDMLR}
\centering 
\scalebox{1.16}{
\begin{tabular}{lcccc}
\toprule
\multirow{2}{*}{\textbf{Method}} & \multicolumn{2}{c}{\textbf{CIFAR-10-LT}} & \multicolumn{2}{c}{\textbf{CIFAR-100-LT}} \\ 
\cmidrule(lr){2-3} \cmidrule(lr){4-5}
 &\textbf{IF=10} &\textbf{IF=100} &\textbf{IF=10} &\textbf{IF=100} \\ 
\midrule
BALMS~\cite{ren2020balanced}          &91.3  &84.9  &63.0  &50.8  \\
LWS~\cite{kang2020decoupling}         &91.1  &83.7  &63.4  &50.5  \\
SSD~\cite{li2021self}                 &-     &-     &62.3  &46.0  \\
t-vMF~\cite{kobayashi2021t}           &91.2  &83.8  &64.7  &50.3  \\
CE+DRS~\cite{zhou2022deep}            &-     &78.78 &-     &45.53 \\
RIDE+CMO~\cite{park2022majority}      &-     &-     &60.2  &50.0 \\
\hdashline

FASA~\cite{zang2021fasa}              &-     &-     &-     &45.2 \\
SAFA~\cite{hong2022safa}              &88.94 &80.48 &59.11 &46.04 \\
\midrule

CE                             &88.22  &72.46 &58.70 &41.28 \\
Label shift~\cite{tian2020posterior_labelshift}  &89.46  &80.88 &61.81 &48.58 \\
WCDAS~\cite{han2023wrapped}    &92.48  &84.67 &65.92 &50.95 \\ 
\hdashline

CE+LDMLR           &89.13 ($\uparrow$0.91)  &76.26 ($\uparrow$3.80)  &60.10 ($\uparrow$1.40) &43.34 ($\uparrow$2.06) \\
Label shift+LDMLR  &89.70 ($\uparrow$0.24)  &82.77 ($\uparrow$1.89) &62.67 ($\uparrow$0.86) &49.76 ($\uparrow$1.18) \\
WCDAS+LDMLR        &\textbf{92.58} ($\uparrow$0.10) &\textbf{86.29} ($\uparrow$1.62) &\textbf{66.32} ($\uparrow$0.40) &\textbf{51.92} ($\uparrow$0.97) \\
\bottomrule
\end{tabular}
}
\end{table*}

\subsection{Setup}\label{sec:setup}
\noindent \textbf{Dataset.}
We evaluate our method on CIFAR-LT~\cite{liu2019large} and ImageNet-LT~\cite{liu2019large} datasets. The CIFAR-LT experiments involve four scenarios: CIFAR-10~\cite{krizhevsky2009learning} with an imbalance factor of $100$, CIFAR-10 with an imbalance factor of $10$, CIFAR-100~\cite{krizhevsky2009learning} with an imbalance factor of $100$, and CIFAR-100 with an imbalance factor of $10$. ImageNet-LT is a subset derived from ImageNet-2012~\cite{ILSVRC15}, containing $1000$ categories. The training set includes $115.8$K images. The number of images per category ranges from $1280$ to $5$. The validation and test sets are balanced datasets containing $20$K and $50$K images, respectively.

\noindent \textbf{Implementation details.}
For CIFAR-LT, before training the diffusion model, a ResNet-32~\cite{he2015deep} is pre-trained with a learning rate of $1e-4$ and a dropout rate of $0.5$ in the first stage of the model training. We then train the diffusion model for $200$ epochs using the Adam optimizer~\cite{kingma2014adam} with a learning rate of $1e-3$ in the second stage. The number of diffusion steps is $1,000$, and $500$ for the reverse steps. The input and output sizes of ResNet-32 are $3\times3\times32$ and $64\times1$, respectively. In the third stage, we reduce the learning rate to $5e-4$ when fine-tuning the fully connected layer of the classifier. For ImageNet-LT, ResNet-10~\cite{he2015deep} and the diffusion model are trained for $200$ epochs, while the final fine-tuning process uses $100$ epochs. The learning rate and optimizer are the same as those used for CIFAR-LT experiments. The batch size is $128$ for all training processes. $\gamma$ is set as $0.05$. We conduct all experiments using an NVIDIA GTX 4080 with $16$~GB Graphic RAM.

\subsection{Results on CIFAR-LT}
Three baselines, i.e. Cross Entropy (CE), label shift~\cite{tian2020posterior_labelshift} and WCDAS~\cite{han2023wrapped}, are selected as baselines for the experimental comparison, and they are trained on the CIFAR-LT dataset~\cite{liu2019large}.
The experimental results are presented in Table~\ref{tab:CIFAR_LDMLR}, from which we can find that the proposed method has improved classification accuracy over the baselines. Notably, the WCDAS+LDMLR method achieves the highest classification accuracy across both datasets and all imbalance factors, with the best performance highlighted in bold. For CIFAR-10-LT with an IF of 100, it reaches an accuracy of $86.29$\% (an improvement of $1.62$\% over the baseline WCDAS method), and for CIFAR-100-LT with an IF of $100$, it achieves an accuracy of $51.92$\% (an improvement of $0.97$\% over the baseline). These results show the effectiveness of combining WCDAS with LDMLR for addressing class imbalance in image classification tasks. It is also observed that our method brings more benefits over the highly imbalanced dataset. For example, on the CIFAR-10-LT, the accuracy gain ($3.80$\%, $1.89$\%, $1.62$\%) with IF 100 for CE, Label shift, and WCDAS are much higher than that ($0.91$\%, $0.24$\%, $0.10$\%) with IF $10$, and the similar improvement is found for the CIFAR-100-LT as well. This helps demonstrate the effectiveness of using feature-generation approaches to tackle the challenges of long-tailed recognition.

\subsection{Results on ImageNet-LT}\label{res_imgnet}
We conduct comparison experiments on the ImageNet-LT dataset~\cite{liu2019large} with the same baselines used on CIFAR-LT. The experimental results are presented in the Table~\ref{tab:ImageNet_LDMLR}. The baselines are first trained to learn the image feature representation and then combined with our method for augmented latent features. The WCDAS+LDMLR method showcases the highest overall accuracy of $44.8$\% among the augmented approaches, indicating a modest improvement of $0.2$\% over the non-augmented WCDAS method. 
Interestingly, the CE+LDMLR method shows a more pronounced overall improvement of $1.4$\%, suggesting that the impact of LDMLR varies with the underlying method. 
It can also be observed from experimental results on the ImageNet-LT dataset that our method is good at improving the classification accuracy for tail classes.

\begin{table*}[t]
\caption{Experiemtal results on ImageNet-LT~\cite{liu2019large}. The encoder is ResNet-10~\cite{he2015deep}. The classification accuracies in $\%$ are provided. ``$\uparrow$'' indicates the improvements over the baseline. The best numbers are in bold.} \label{tab:ImageNet_LDMLR}
\centering
\resizebox{0.6\textwidth}{!}{
\begin{tabular}{lcccc}
\toprule
\multirow{2}{*}{\textbf{Method}}                    &\multicolumn{4}{c}{\textbf{ImageNet-LT}} \\ \cmidrule(lr){2-5}
    &\textbf{Many} &\textbf{Medium} &\textbf{Few} &\textbf{All} \\ \midrule

cRT~\cite{kang2020decoupling}   &49.9 &37.5 &23.0 &40.3 \\
LWS~\cite{kang2020decoupling}   &48.0 &37.5 &22.8 &39.6 \\
BALMS~\cite{ren2020balanced}    &48.0 &38.3 &22.9 &39.9 \\
t-vMF~\cite{kobayashi2021t}     &55.4 &39.9 &22.5 &43.5 \\
\midrule

CE                                &57.7 &26.6 &4.4 &35.8 \\
Label shift~\cite{tian2020posterior_labelshift} &52.0 &39.3 &20.3 &41.7 \\
WCDAS~\cite{han2023wrapped}       &57.1 &40.9 &23.3 &44.6 \\ 
\hdashline

CE+LDMLR              &57.2 &29.2 &7.3 &37.2 ($\uparrow$1.4) \\
Label shift+LDMLR     &50.9 &39.4 &23.7 &42.2 ($\uparrow$0.5) \\
WCDAS+LDMLR           &57.0 &41.2 &23.4 &\textbf{44.8} ($\uparrow$0.2) \\
\bottomrule
\end{tabular}}
\end{table*}

\begin{table*}[t]
\caption{Ablation study: augmentation on the image level. The classification accuracies in $\%$ are provided. The best numbers are in bold. The CE+DM and Lable shift+DM denote that the diffusion model is applied to generate image-level data for augmentation.}
\label{tab:CIFAR_DM}
\centering
\resizebox{0.65\textwidth}{!}{
\begin{tabular}{lcccc}
\toprule
\multirow{2}{*}{\textbf{Method}}& \multicolumn{2}{c}{\textbf{CIFAR-10-LT}} & \multicolumn{2}{c}{\textbf{CIFAR-100-LT}} \\ 
\cmidrule(lr){2-3} \cmidrule(lr){4-5}
 &\textbf{IF=10} &\textbf{IF=100} &\textbf{IF=10} &\textbf{IF=100} \\ 
\midrule
CE                                                 &88.22 & 72.46 & 58.70 & 41.28 \\
Label shift~\cite{tian2020posterior_labelshift}    &89.46 & 80.88 & 61.81 & 48.58 \\ 
\hdashline
(Image level) \\
CE+DM           &88.88  &73.91  &59.19   &42.41  \\
Label shift+DM  &89.63  &82.10  &61.96   &48.93  \\ 
\hdashline
(Feature level) \\
CE+LDMLR (Ours)          &\textbf{89.13}  &\textbf{76.26}  &\textbf{60.10}  &\textbf{43.34}  \\
Label shift+LDMLR (Ours) &\textbf{89.70}  &\textbf{82.77}  &\textbf{62.67}  &\textbf{49.76}  \\
\bottomrule
\end{tabular}}
\end{table*}

\subsection{Analysis}\label{sec:ablation_study}
\noindent \textbf{Augmentation on the image level.}\label{sec:img_aug}
We study the effectiveness of image augmentation using a diffusion model and demonstrate the importance of data augmentation in latent space. To accomplish this, we train a conditional diffusion model on CIFAR-LT and use it to generate new images. These generated images are combined with the original long-tailed data to create a new image dataset, and we examine the accuracy with and without these augmented images. As presented in Table ~\ref{tab:CIFAR_DM}, feature-level augmentation (CE+LDMLR and Label shift+LDMLR) consistently outperforms image-level augmentation (CE+DM and Label shift+DM) across all settings. Specifically, in the CIFAR-10-LT dataset with IF of $100$, the Label shift+LDMLR method achieves the highest classification accuracy of $82.77$\%, demonstrating a significant improvement over the Label shift+DM method with $82.10$\%. Similarly, in the CIFAR-100-LT dataset with IF of $100$, the Label shift+LDMLR method also records the highest accuracy of $49.76$\%, surpassing the Label shift+DM method ($48.93$\%). 
The limited accuracy gain of image-level augmentation could be caused by the difficulty in generating high-fidelity image samples from a limited-scale training set. The feature representation in a latent space with lower dimensions might be more easily learned than that in the image space.

\begin{figure}[t]
  \centering
   \includegraphics[width=1.0\linewidth]{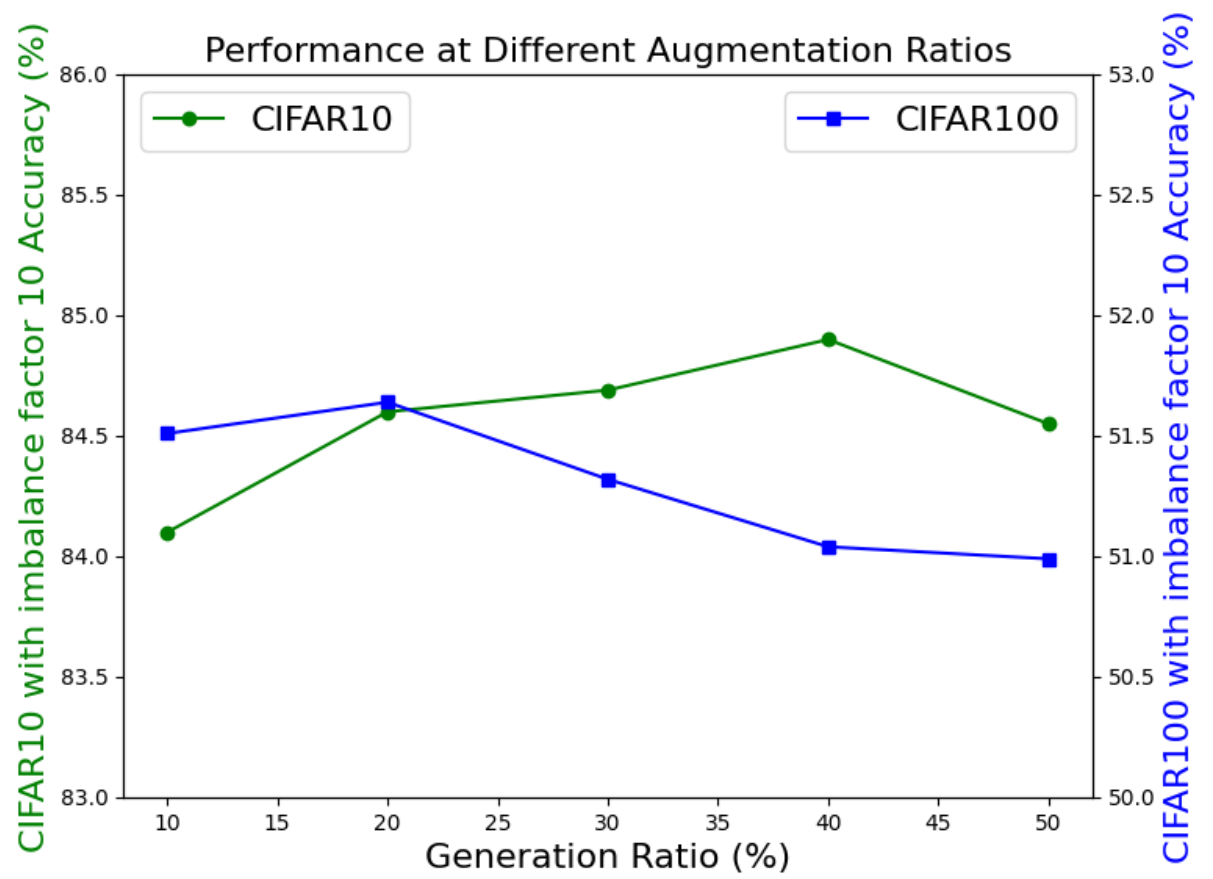}
   \caption{The impact of generation ratio on classification accuracy. The evaluation is conducted on CIFAR-10-LT and CIFAR-100-LT with $\mathrm{IF}=10$.}
   \label{fig:feature_vis}
\end{figure}

\noindent \textbf{Augmentation ratio.}\label{sec:aug_ratio}
The number of generated features is important to the performance. The augmentation ratio represents the proportion between the generated and the encoded features, and we investigate the impact of this ratio on dataset CIFAR-10-LT and CIFAR-100-LT with IF of $10$, as shown in Figure~\ref{fig:feature_vis}. As for CIFAR-10-LT, the classification accuracy degrades when the generation ratio is over $20$\%, while for CIDAR-100-LT, the accuracy goes up with the generation ratio until it passes $40$\%. This difference could be caused by the smaller number of tail classes in CIFAR-100-LT than in CIFAR-10-LT. 

\noindent \textbf{Effects of tail category: many/medium/few.}\label{sec:mmf}
Here, we investigate the effect of augmenting features from different class distributions—many, medium, few—on CIFAR-LT when $\mathrm{IF}=100$, as shown in Table~\ref{tab:many-medium-few}.  We compare the baseline method, which does not use any augmented features, with strategies that augment features for all classes and selectively for many, medium, or few classes. For both settings, augmenting features for ``few'' classes consistently yields the highest classification accuracy, highlighting the effectiveness of focusing augmentation efforts on underrepresented classes. Specifically, on CIFAR-10-LT, the WCDAS+LDMLR method with augmentation on ``few'' classes achieves a top accuracy of $86.29$\%, demonstrating a significant improvement over the baseline accuracy of $84.67$\%. Similarly, on CIFAR-100-LT, the same method and augmentation strategy lead to the best accuracy of $51.92$\%, compared to the baseline's $50.95$\%. These results highlight the performance of targeted feature augmentation in generating ``few'' classes. However, this is specific to CIFAR-LT, and we augment ``all'' classes for ImageNet-LT.

\begin{table*}[t]
\caption{Ablation study: many/medium/few. The classification accuracies in $\%$ are provided. The best numbers are in bold. ``All'', ``Many'', ``Medium'' and ``Few'' indicate augmenting features of ``all'', ``many'', ``medium'' and ``few'' classes, respectively.}
\label{tab:many-medium-few}
\centering
\resizebox{1.0\textwidth}{!}{
\begin{tabular}{lccccc|ccccc}
\toprule
\multirow{2}{*}{\textbf{Method}}& \multicolumn{5}{c}{\textbf{CIFAR-10 -LT (IF=100)}} & \multicolumn{5}{|c}{\textbf{CIFAR-100-LT (IF=100)}} \\ 
\cmidrule(lr){2-6} \cmidrule(lr){7-11}
 &\textbf{Baseline} &\textbf{``All''} &\textbf{``Many''} &\textbf{``Medium''} &\textbf{``Few''} &\textbf{Baseline} &\textbf{``All''} &\textbf{``Many''} &\textbf{``Medium''} &\textbf{``Few''} \\ 
\midrule
CE+LDMLR                        &72.46 &76.26 &74.03 &74.85 &76.26 &41.28 &42.82 &41.22 &42.79 &43.34  \\
Label shift+LDMLR               &80.88 &81.24 &78.22 &78.41 &82.77 &48.58 &48.71 &44.57 &48.71 &49.76  \\
WCDAS+LDMLR                     &84.67 &84.90 &83.51 &83.62 &86.29 &50.95 &51.64 &49.38 &51.64 &51.92 \\
\bottomrule
\end{tabular}}
\end{table*}

\noindent \textbf{Impact of encoded and generated features.} This ablation study focuses on the impact of encoded and generated features on classification accuracy, as shown in Table~\ref{tab:encoded and generated features}. The result encompasses experiments across two datasets: CIFAR-10-LT and CIFAR-100-LT, with IF of $10$ and $100$. The comparison includes approaches utilizing only encoded features, augmentations with generated features, and combining both encoded and generated features. These results demonstrate the effectiveness of the combination of encoded and generated features in addressing the long-tailed classification problems.

\begin{table*}[t]
\caption{Ablation study: encoded and generated features. The classification accuracies in $\%$ are provided. The best numbers are in bold.}
\label{tab:encoded and generated features}
\centering
\resizebox{0.86\textwidth}{!}{
\setlength{\tabcolsep}{18pt} 
\begin{tabular}{lcccc}
\toprule
\multirow{2}{*}{\textbf{Method}}& \multicolumn{2}{c}{\textbf{CIFAR-10 -LT}} & \multicolumn{2}{c}{\textbf{CIFAR-100-LT}} \\ 
\cmidrule(lr){2-3} \cmidrule(lr){4-5}
 &\textbf{IF=10} &\textbf{IF=100} &\textbf{IF=10} &\textbf{IF=100} \\ 
\midrule
(Only encoded features) \\
Label shift~\cite{tian2020posterior_labelshift}  &89.46  &80.88 &61.81 &48.58 \\
WCDAS~\cite{han2023wrapped}    &92.48  &84.67 &65.92 &50.95 \\ 
\hdashline

(Only generated features) \\
Label shift+LDMLR  &89.43    &82.18    &54.52    &32.15 \\
WCDAS+LDMLR        &91.98    &83.83    &64.93    &50.42 \\
\hdashline

(Both features) \\
Label shift+LDMLR  &\textbf{89.70} &\textbf{82.77} &\textbf{62.67} &\textbf{49.76} \\
WCDAS+LDMLR        &\textbf{92.58} &\textbf{86.29} &\textbf{66.32} &\textbf{51.92} \\
\bottomrule
\end{tabular}}
\end{table*}

\begin{figure}[t]
  \centering
   \includegraphics[width=1.0\linewidth]{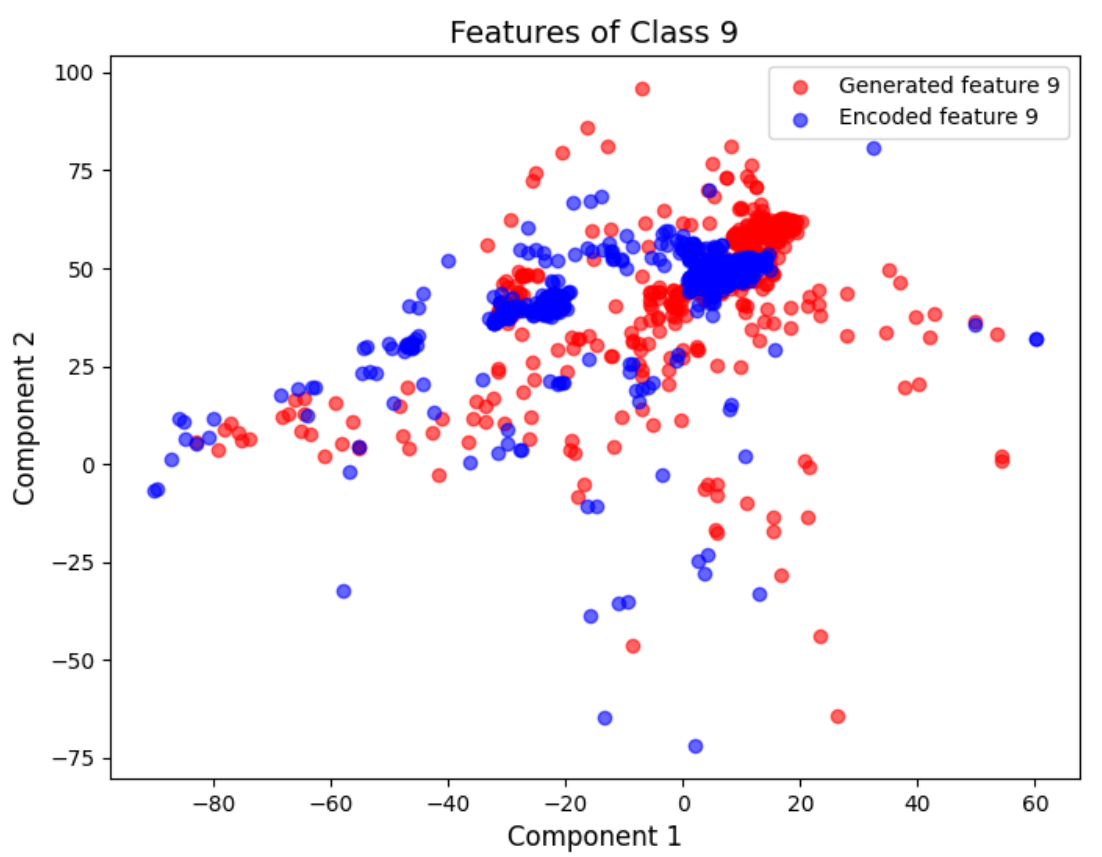}
   \caption{The encoded and generated features of tail class (class 9) in CIFAR-10-LT during the model training. From the figure, the generated features (blue points) can overlay the encoded features (red points) from the original training dataset while slightly enriching the feature space.}
   \label{fig:feature_vis2}
\end{figure}

\noindent \textbf{Visualization of generated features} In Figure~\ref{fig:feature_vis2}, we visualize feature embeddings during the model training. The lower figure shows the encoded and generative feature distributions of the tail class for CIFAR-10 with an imbalance factor of $0.1$. By comparing the distribution of encoded features and those generated by the diffusion model, we observe that the generated features can overlap with parts of the distribution of the encoded features while moderately enriching the original distribution, thereby achieving the goal of feature augmentation effectively. 

\noindent \textbf{Future works.} The training of our LDMLR requires multiple stages. Hence, one future work could be the simplification of its training process. Another future work could be the exploration of more diffusion models. Lastly, the quality of feature augmentation depends on the diffusion model's generation quality on long-tailed distributed data. Therefore, enhancing the quality of feature augmentation depending on the generation of the diffusion model on long-tailed distributed datasets might be an important future task.
\section{Conclusion}
This work proposes a novel framework, LDMLR, to address the challenge of long-tailed recognition. The LDMLR leverages the powerful generative capabilities of diffusion models for latent-level data augmentation, aiming to balance long-tailed distributed datasets. To the best of our knowledge, we are the first to adopt the diffusion model in the long-tailed problems. The experimental outcomes show our method has improvements in several datasets. We hope that this work could motivate more practical uses of the diffusion model.

{
 \small
 \bibliographystyle{ieeenat_fullname}
 \bibliography{main}
}

\end{document}